\pdfoutput=1

\documentclass[11pt]{article}

\usepackage[]{acl}

\usepackage{float}
\usepackage{times}
\usepackage{latexsym}
\usepackage{amsmath}
\usepackage{amssymb}
\usepackage{arydshln}
\usepackage{pgfplots}
\usepgfplotslibrary{fillbetween} 
\usepackage{subcaption}
\usepackage{algorithm}
\usepackage{algpseudocode}
\usepackage[T1]{fontenc}
\usepackage{mathtools}
\usepackage{svg}
\usepackage{enumitem}
\usepackage{tabu}
\usepackage{tabularx}
\usepackage{multirow}
\usepackage{booktabs}
\usepackage{color}
\usepackage{xcolor}
\usepackage{colortbl}
\usepackage[noabbrev,capitalise]{cleveref}
\usepackage{pifont}
\usepackage[utf8]{inputenc}
\usepackage{microtype}
\usepackage{bm,bbm}
\def\vtheta{{\bm{\theta}}}
\def\vlambda{{\bm{\lambda}}}
\def\vtau{{\bm{\tau}}}
\def\vdelta{{\bm{\delta}}}
\def\vx{{\mathbf{x}}}
\def\vy{{\mathbf{y}}}
\def\vf{{\mathbf{f}}}
\def\mF{{\mathbf{F}}}
\newcommand{\vthetaFT}{{\vtheta_{\text{ft}}}}
\newcommand\vthetaPRE[0]{{{\vtheta_0}}}
\newcommand\vthetaNEW[0]{{\vtheta^\prime}}
\newcommand\vthetaE[0]{{\vtheta_{\text{E}}}}
\newcommand\vthetaAE[0]{{\vtheta_{\text{AE}}}}

\newcommand{\improv}[1]{\scriptsize{\color{PineGreen}($\downarrow$#1)}}
\newcommand{\improvUP}[1]{\scriptsize{\color{PineGreen}($\uparrow$#1)}}

\newcommand{\degrad}[1]{\scriptsize{\color{BrickRed}($\downarrow$#1)}}

\newcommand{\degradUP}[1]{\scriptsize{\color{BrickRed}($\uparrow$#1)}}
\newcommand{\nochange}[1]{\scriptsize{\color{black}(-#1)}}
\newcommand{\finetune}[2]{\scriptsize{\text{finetune}(#1, #2)}}

\usepackage{todonotes}
\makeatletter
\newcommand*\iftodonotes{\if@todonotes@disabled\expandafter\@secondoftwo\else\expandafter\@firstoftwo\fi}
\makeatother

\definecolor{edolime}{rgb}{0.9,1,0.3}

\newcommand{\cmark}{\color{ForestGreen}\ding{51}}
\newcommand{\xmark}{\color{BrickRed}\ding{55}}

\newcolumntype{L}[1]{>{\raggedright\let\newline\\\arraybackslash\hspace{0pt}}p{#1}}
\usepackage{xpatch}
\usepackage{tikz}
\usepackage{colortbl}
\usepackage{fontawesome}
\usepackage{contour}
\usetikzlibrary{arrows}

\makeatletter
\xpatchcmd{\algorithmic}
  {\ALG@tlm\z@}{\leftmargin\z@\ALG@tlm\z@}
  {}{}
\makeatother
\title{Elastic Weight Removal for Faithful and Abstractive Dialogue Generation}

\author{Nico Daheim$^1$~~~Nouha Dziri$^2$~~~Mrinmaya Sachan$^3$ \\\textbf{Iryna Gurevych$^1$}~~~\textbf{Edoardo M. Ponti$^4$} \\
$^1$Ubiquitous Knowledge Processing Lab (UKP Lab), Department of Computer Science\\ and Hessian Center for AI (hessian.AI), TU Darmstadt\\~~~$^2$Allen Institute for Artificial Intelligence~~~$^3$ETH Zürich~~~$^4$University of Edinburgh \\
  {\url{www.ukp.tu-darmstadt.de}} }%

\begin{document}
\maketitle
\begin{abstract}
Ideally, dialogue systems should generate responses that are faithful to the knowledge contained in relevant documents. However, many models generate hallucinated responses instead that contradict it or contain unverifiable information. To mitigate such undesirable behaviour, it has been proposed to fine-tune a `negative expert' on negative examples and subtract its parameters from those of a pre-trained model. However, intuitively, this does not take into account that some parameters are more responsible than others in causing hallucinations. Thus, we propose to weigh their individual importance via (an approximation of) the Fisher Information matrix, which measures the uncertainty of their estimate. We call this method Elastic Weight Removal (EWR).
We evaluate our method---using different variants of Flan-T5 as a backbone language model---on multiple datasets for information-seeking dialogue generation %
and compare our method with state-of-the-art techniques for faithfulness, such as CTRL, Quark, DExperts, and Noisy Channel reranking.
Extensive automatic and human evaluation shows that EWR systematically increases faithfulness at minor costs in terms of other metrics. %
However, we notice that only discouraging hallucinations may increase extractiveness, i.e. shallow copy-pasting of document spans, which can be undesirable. 
Hence, as a second main contribution, we show that our method can be extended to simultaneously discourage hallucinations \emph{and} extractive responses. %
We publicly release the code for reproducing EWR and all %
baselines.
\end{abstract}

\hspace{.5em}\includegraphics[width=1.25em,height=1.25em]{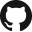}\hspace{.75em}\parbox{\dimexpr\linewidth-2\fboxsep-2\fboxrule}{\url{https://github.com/ndaheim/faithful-dialogue}}
\vspace{-.5em}
\section{Introduction}

Current-day large language models (LLMs) habitually generate coherent, grammatical and seemingly meaningful sentences of text. The rise of dialogue language models, most notably ChatGPT \cite{team2022chatgpt}, GPT4 \cite{openai2023gpt4} and LaMDA \cite{thoppilan2022lamda}, has revolutionised the field of natural language processing (NLP).
            \begin{figure}[t!]
    \centering
    \begin{tabularx}{\columnwidth}{Xcc}
       \multicolumn{3}{X}{$\mathcal{K}$: The Flash first appeared in ``Showcase'' \#4 (October 1956) [...]} \\
       \multicolumn{3}{X}{$u_T$: What comic series is he from?} \\
       \hline
       $u_{T+1}$ & F & A \\
        He \textit{first appeared in ``Showcase'' \#4} (\textcolor{red}{November} 1956). & {\xmark} & {\xmark} \\
        He \textit{first appeared in ``Showcase'' \#4 (October 1956)}. & {\cmark} & {\xmark} \\
        His first appearance was \textit{in Showcase \#4} in October 1956. & {\cmark} & {\cmark}
    \end{tabularx}
    \caption{Constructed example of responses $u_{T+1}$ that are i) hallucinated (words contradicting the knowledge $\mathcal{K}$ in red); ii) faithful but not abstractive (longest copied $n$-gram in italic); and iii) both \textbf{F}aithful and \textbf{A}bstractive based on Wizard-of-Wikipedia \cite{dinan2018wizard}.}
    \label{fig:figure1}
\end{figure}

However, despite their impressive capabilities, these systems sometimes hallucinate,\footnote{We define hallucination as the antonym of faithfulness.} fail to adhere to the ground-truth knowledge, and do not specify their degree of confidence \citep[\textit{inter alia}]{wiseman-etal-2017-challenges,dziri2022faithdial,ji2022survey}. 
Hallucinations severely limit their suitability and raise safety concerns. 
For instance, they may impair student learning, proliferate convincing-but-inaccurate news articles, and result in life-threatening medical decisions. 
Therefore, ensuring that these systems are trustworthy is crucial for their safe deployment at scale, particularly in high-stakes domains.

Modelling solutions to mitigate hallucination can take inspiration from methods devised to discourage other undesirable behaviours in LLMs, such as contradictions \citep{keskar2019ctrl}, repetitions \citep{lu2022quark}, or toxicity \citep{ilharco2023editing}. 
A group of methods achieves this goal by fine-tuning an LLM to be conditioned on special tokens \citep{keskar2019ctrl} or to generate samples with high scores according to a learned reward function \citep{lu2022quark}. 
Another re-weights the predictive distribution with experts and anti-experts \citep{liu2021dexperts, daheim-etal-2022-controllable}.
Here, expert denotes a model that is trained to perform desirable behaviour, while anti-expert denotes a model trained to perform undesirable behaviour.
While successful, these methods are either time inefficient as they require sampling a large number of generations during training, or space inefficient as they store and evaluate multiple models during inference.
On the other hand, a third family of methods
\citep{choubey2021cape,ilharco2023editing} has proposed to unlearn negative behaviours by interpolating parameters, which achieves better efficiency \textit{without altering the model architecture} by using modular deep learning \citep{ponti-etal-2021-parameter,pfeiffer2023modular}. Specifically, a new model is obtained as the weighted difference between a pre-trained LLM and a \emph{task vector} --obtained by subtracting a finetuned version from the pretrained LLM-- which is known as \textit{task arithmetic}.
However, this does not consider that the degree of responsibility for a certain behaviour might vary across individual parameters.
Secondly, it might result in catastrophic interference between models specialised for different properties \citep{mccloskey1989catastrophicinterference}.

In order to address these issues, we propose Elastic Weight Removal (EWR), a novel method that performs parameter interpolation while weighting the importance of each parameter to \textit{remove undesired behaviours}.
In particular, we show how this idea can be applied to discouraging hallucinated generations in dialogue.
First, we train an anti-expert on a synthetically created dataset.
Then, we interpolate this anti-expert with the baseline model using their respective Fisher Information Matrix (FIM) as a measure of importance, which represents the precision of their estimates.
This is reminiscent of previous uses of the FIM, such as continual learning \citep{kirkpatrick2017overcoming}, sample-efficient learning \citep{ponti2019towards}, or merging models for different tasks \citep{matena2021merging}.
However, we show that faithfulness may have the side effect of increased extractiveness, i.e., copy-pasting passages of the document(s) with a shallow understanding of their meaning. 
Hence, we try to improve both faithfulness \textit{and} abstractiveness, which corresponds to re-phrasing the ground-truth knowledge and drawing new valid conclusions from it, by not only interpolating with a hallucination anti-expert but also an abstractiveness expert trained on a subset of our training data that is categorised as "abstractive" according to \textit{n}-gram overlap.

We assess the effectiveness of our approach on various information-seeking dialogue benchmarks, namely WoW \citep{dinan2018wizard}, DSTC9 and DSTC11 \citep{kim2020beyond}, and FaithDial \citep{dziri2022faithdial}. 
We compare our method with state-of-the-art techniques for unlearning negative behaviours that we \textit{adapt for removing hallucinations}, including CTRL \citep{keskar2019ctrl}, Quark \citep{lu2022quark}, DExperts \citep{liu2021dexperts}, Noisy Channel reranking \citep{daheim-etal-2022-controllable}, and task arithmetic \citep{choubey2021cape,ilharco2023editing}. 
Our findings reveal that EWR consistently enhances faithfulness with minor impact on the abstractiveness of the responses. Moreover, EWR can be integrated with other methods like CTRL to achieve state-of-the-art results. 
We confirm our findings with a human evaluation with expert annotators based on the Attributable to Identified Source (AIS) framework  \cite{rashkin2021measuring}, which highlights gains of EWR in the `Attributable' and `Paraphrasing' dimensions. 
These correspond to faithful and abstractive generations, respectively. 
To summarise, our main contributions are:
\begin{itemize}[noitemsep,nolistsep,leftmargin=11pt]
    \item A novel method for \emph{importance weighted task arithmetic} based on Fisher Information (\S\ref{sec:ewr}).
    \item A framework for removing undesired behaviours in dialogue \textit{along multiple dimensions}. Specifically, we rely on simple metrics and data augmentations to create and identify hallucinated and extractive examples (\S\ref{sec:expert_anti_expert}).
    \item The adaptation of a series of techniques devised for unlearning undesired behaviours to faithful dialogue generation (\S\ref{sec:baselines}). We offer the first comparison among them in a controlled setting, with the same negative examples and the same LLM backbone (\S\ref{sec:results}). 
    \item A comprehensive open-source repository for reproducing EWR, baselines, datasets, and metrics contained in the present paper: \url{https://github.com/ndaheim/faithful-dialogue}.
\end{itemize}
\section{Background}
\label{sec:background}
The goal of response generation in dialogue is to generate a system turn $u_{T+1}$ as a continuation of a dialogue $u_1^T \coloneqq (u_1, \dots, _T)$ consisting of $T$ turns that are usually taken by either the system or a user that interacts with it.
Each dialogue turn $u_t$ is a sequence of $N_t$ tokens $[u_t]_1^{N_t} \in \mathcal{V}^{N_t}$ from the model vocabulary $\mathcal{V}$. 
In document-grounded response generation, $u_{T+1}$ is grounded in one or more documents $\hat{\mathcal{K}} \subseteq \mathcal{K}$ from a \emph{document knowledge base} $\mathcal{K}$. The model is conditioned on $\hat{\mathcal{K}}$, which gives relevant information for the next turn.
Crucially, $u_{T+1}$ generated by the model should faithfully reflect the knowledge contained in $\hat{\mathcal{K}}$, not contradict their information content, nor add unverifiable information.
Similar to previous work in this area, we do not model the retrieval of grounding documents; rather, we assume that the annotated \emph{ground-truth} documents are available in an oracle setting, to focus our study on faithful response generation.

More specifically, we study locally-normalised neural language generators of the form: \begin{align}
    &p_\vtheta(u_{T+1} \mid u_1^T, \hat{\mathcal{K}}) = \nonumber \\
    &\quad \quad \prod_{n=1}^{N_{T+1}} p_\vtheta([u_{T+1}]_{n} \mid [u_{T+1}]_{1}^{n-1}, u_1^T, \hat{\mathcal{K}}) \text{,}
\end{align}
which are parameterised by the weights $\vtheta$. Specifically, we wish to find $\vtheta$ such that it increases performance in dialogue generation while retaining faithfulness and abstractiveness. The second property ensures that the resulting model does not simply collapse into parroting the documents $\hat{\mathcal{K}}$ but rather grasps their meanings and paraphrases them. We focus on different methods of obtaining $\vtheta$ \emph{without altering} the LLM architecture.

We first introduce a general formulation of obtaining $\vtheta$ by combining the parameters of multiple models in \cref{sec:param_combination} before going over different concrete merging functions.
Then, in \cref{ssec:fisher}, we define Fisher Information as a measure of the importance of individual parameters. This constitutes a crucial ingredient of our proposed recipe for removing hallucination and extractiveness---Elastic Weight Removal---that is finally introduced in \cref{sec:ewr}.

\subsection{Parameter Combination}
\label{sec:param_combination}
Previous works have already explored methods to \emph{combine model parameters}: for example, averaging checkpoints to increase robustness \cite{gao2022revisiting} or promoting desirable behaviours by merging specifically trained model instances \cite{ilharco2023editing}.
By letting $\Theta = \{\vtheta_1, \dots, \vtheta_N\}$, where $\vtheta_i \in \mathbb{R}^d$, denote the parameters of a set of models that should be merged and $\vlambda_i \in \mathbb{R}^d$ their respective scaling factors, we may introduce a general formulation of parameter combination as a weighted sum, given a normalisation constant $Z$:
 \begin{equation}
 \label{eq:param_combination}
     \vthetaNEW = \sum_{i=1}^N \frac{\vlambda_i \odot \vtheta_i}{Z} \text{.}
 \end{equation}
where $\odot$ denotes the Hadamard product.
In what follows, we discuss different instantiations of this function, which were used to \emph{steer} the model behaviour.

\subsubsection{Task Arithmetic}
The core idea of task arithmetic is that essential information about a task can be captured by the change of the parameter values between pretrained initialisation $\vthetaPRE$
and after task fine-tuning $\vthetaFT$, which we call \textit{task vector}. \citet{ansell-etal-2022-composable} showed that (sparse) task vectors from multiple tasks can be composed by addition.\footnote{In terms of \cref{eq:param_combination}, this amounts to setting $Z = 1$ and $\vlambda \in \{0, 1\}^d$.} \citet{ilharco2023editing} extended this framework to further arithmetic operations. 
Task vectors can not only promote a specific behaviour by addition but also  suppress it by subtraction. 
Concretely, we can make use of the latter to negate hallucinations by subtracting the parameters of a model $\vthetaAE$ trained on hallucinated examples. 
AE refers to \emph{anti-expert} and we use E to index an \emph{expert} model.

Formally, the task vector $\vtau$ may be described as \begin{equation}
\label{eq:task_vector}
\vtau \coloneqq \vthetaAE - \vthetaPRE \text{ .}
\end{equation}
We obtain the function to compose the pretrained parameters $\vthetaPRE$ with the task vector $\vtau$, as proposed by \citet{ilharco2023editing}, from \cref{eq:param_combination} by setting $Z=1$, $\vlambda_{\vthetaPRE} = \bm{1}$ and $\vlambda_{\vtau} = \lambda\bm{1}$. Crucially, AE task vectors are \emph{subtracted}, whereas E task vectors are \textit{added}. This results in the following equation: 
\begin{align}
    \vthetaNEW &= \vthetaPRE - \lambda \cdot \vtau \nonumber \\
    &= \vthetaPRE - \lambda\cdot \vthetaAE + \lambda \cdot \vthetaPRE \nonumber \\
    &= (1+\lambda) \cdot \vthetaPRE - \lambda \cdot \vthetaAE \text{.}
    \label{eq:task_arithmetic}
\end{align}
Under this lens, task arithmetic can be generalised to the combination of an arbitrary number of task vectors as:
\begin{align}
\label{eq:task_arithmetic_general}
    \vthetaNEW = \vthetaPRE + \sum_{i} \lambda_i \vtau_i \text{,}
\end{align}
where $\lambda_i > 0$ implies promotion and $\lambda_i < 0$ suppression.

By restricting Equation \ref{eq:task_arithmetic_general} to just using  one anti-expert $\vthetaAE$ and one expert $\vthetaE$ with weights $\lambda \coloneqq \lambda_1 = \lambda_2$ and task vectors $\vtau_1$ and $\vtau_2$, we can recover Contrastive Parameter Estimation \citep[CaPE;][]{choubey2021cape}:\footnote{In CaPE, the models are trained on disjoint partitions of the training data.
However, this restriction may be lifted to use augmented data or the same data point in both partitions.} 
\begin{align}
        \vthetaNEW &= \vthetaPRE -\lambda \cdot \vtau_1 + \lambda\cdot \vtau_2 \nonumber \\
        &= \vthetaPRE - \lambda\cdot \vthetaAE + \lambda \cdot \vthetaPRE + \lambda\cdot \vthetaE - \lambda \cdot \vthetaPRE \nonumber \\
        &= \vthetaPRE - \lambda\cdot \vthetaAE + \lambda  \cdot \vthetaE \nonumber \\
        &= \vthetaPRE + \lambda\cdot( \vthetaE -\vthetaAE ) \label{eq:cape}
\end{align}
Furthermore, noticing the similarity between \cref{eq:cape} and \cref{eq:task_arithmetic}, one might initialise $\vthetaPRE$ by setting $\vthetaPRE = \vthetaE$, such that task arithmetic is performed on the expert model directly rather than on the pretrained model.

However, both task arithmetic and CaPE assume equal parameter importance, since the scaling factor is identical for all parameters.
One might question this assumption, as potentially only a subset of parameters induces hallucinations.
For example, anomalous encoder--decoder attention patterns correlate strongly with hallucinations \citep[\textit{inter alia}]{raunak-etal-2021-curious, guerreiro2022optimal}. Hence, only these specific parameters might be required to change. %
Moreover, these methods might not be suited to composing multiple task vectors, similarly to \cref{eq:task_arithmetic_general}: in fact, \citet{ansell-etal-2022-composable} showed that this may lead to catastrophic interference.

To address these limitations, we take inspiration from a series of prior works using parameter-specific scaling, for example for mitigating catastrophic forgetting \cite{kirkpatrick2017overcoming}, merging checkpoints of the same model trained independently on different tasks \cite{matena2021merging}, or using sparse masks for fine-tuning \cite{guo2021parameter, sung2021training,ansell-etal-2022-composable}.
In the following, we first introduce Fisher Information as one way of obtaining parameter-specific scaling before introducing our main contribution, Elastic Weight Removal, in \cref{sec:ewr}.

\subsection{Fisher Information}
\label{ssec:fisher}
Given a model $p_{\vtheta}(\vy \mid \vx)$ that induces a conditional distribution over $\vy$ given $\vx$ and is parameterised by $\vtheta$, the Fisher Information matrix $\mF_\vtheta$, commonly referred to as \emph{Fisher}, is defined as the covariance of its score function  $\nabla_\vtheta \log p_\vtheta(y\mid x)$: \begin{align}
    \mF_{\vtheta} = \mathbb{E}_{p_\vtheta(\vy\mid \vx)} \nabla_\vtheta \log p_\vtheta(\vy\mid \vx) \nabla_\vtheta \log p_\vtheta(\vy\mid \vx)^\top
\end{align}
Since the expectation is oftentimes intractable to compute exactly, the Fisher is commonly approximated as the \emph{empirical} or \emph{observed} Fisher
\begin{align}
    \mF_{\vtheta} \approx \frac{1}{|\mathcal{D}|} \sum_{\mathcal{D}} \nabla_\vtheta \log p_\vtheta(\vy\mid \vx) \nabla_\vtheta \log p_\vtheta(\vy\mid \vx)^\top \text{.}
\end{align}
Not only that, as the size of $\mF_{\vtheta}$ scales quadratically in $\vtheta$, it is often approximated by its diagonal
\begin{align} \label{eq:diagfisher}
    \vf_{\vtheta} = \frac{1}{|\mathcal{D}|} \sum_{\mathcal{D}} (\nabla_\vtheta \log p(\vy\mid \vx))^2 \text{.}
\end{align}
While the estimator in \cref{eq:diagfisher} is unbiased, the square can also be calculated over a mini-batch instead of a single example, which 
is frequently used in stochastic optimisation techniques \cite{kim2022fisher},
for instance to add second-order information \cite{amari98natural}.

In this work, we take advantage of one specific property of the Fisher, namely that it relates changes in the output distribution to a change in parameters.
Following \citet{pascanu2013revisiting}, for a given arbitrarily small $\vdelta$, the divergence in output distribution between the original parameter set $\vtheta$ and a perturbed set $\vtheta + \vdelta$ may be rewritten---based on the second order Taylor approximation---as:
 \begin{align*}
    D_{KL}(p_\vtheta \mid \mid p_{\vtheta + \vdelta}) &\approx \frac{1}{2}\vdelta^\top \mF_\vtheta \vdelta
\end{align*}
as $\vdelta \rightarrow 0$.
This naturally gives rise to the interpretation that the Fisher assigns to each parameter an `importance' proportional to its contribution to a change in prediction under slight perturbation.

In the following, we use this property to motivate our main contribution.
\section{Elastic Weight Removal}
\label{sec:ewr}

\begin{algorithm}
\footnotesize
\textbf{Input} Dialogues $\mathcal{D}$, \textcolor{brown}{hallucinated anti-expert dataset $\mathcal{D}^{\text{AE}}$}, \textcolor{teal}{abstractive expert dataset $\mathcal{D}^{\text{E}}$}, initial parameter set $\vthetaPRE$ \newline
\textbf{Output} $\vthetaNEW$
\begin{algorithmic}
\State $\vthetaPRE \hspace{0.06in} \leftarrow \finetune{\vtheta}{\mathcal{D}}$
\State \textcolor{brown}{$\vthetaAE \leftarrow \finetune{\vthetaPRE}{\mathcal{D}^{\text{AE}}}$}
\State \textcolor{brown}{$\vtau_1 \hspace{0.065in}\leftarrow \vthetaAE - \vthetaPRE$}
\State \textcolor{teal}{$\vthetaE \hspace{0.06in}\leftarrow \finetune{\vthetaPRE}{\mathcal{D}^{\text{E}}}$}
\State \textcolor{teal}{$\vtau_2 \hspace{0.065in}\leftarrow \vthetaE - \vthetaPRE$}
\State {\color{black}$\displaystyle{\vf_{\vthetaPRE}} \hspace{0.03in}\leftarrow \frac{1}{|\mathcal{D}|} \sum_{\mathcal{D}} (\nabla \log p_{\vthetaPRE}(u_{T+1} \mid u_1^T, \hat{\mathcal{K}}))^2$}
\State {\color{brown} $\displaystyle{\vf_{\vtau_1}} \hspace{0.03in} \leftarrow \frac{1}{|\mathcal{D}^{\text{AE}}|} \sum_{\mathcal{D}^{\text{AE}}} (\nabla \log p_{\vtau_1}(u_{T+1} \mid u_1^T, \hat{\mathcal{K}}))^2$}

\State {\color{teal} $\displaystyle {\vf_{\vtau_2}}  \hspace{0.03in}\leftarrow \frac{1}{|\mathcal{D}^{\text{E}}|} \sum_{\mathcal{D}^{\text{E}}} (\nabla \log p_{\vtau_2}(u_{T+1} \mid u_1^T, \hat{\mathcal{K}}))^2$}

\State $\vthetaNEW \hspace{0.07in} \leftarrow \frac{\color{black}{\lambda_0 \cdot \vf_{\vthetaPRE}} \color{black}{\cdot \vthetaPRE}\textcolor{brown}{ -\lambda_1 \cdot \vf_{\vtau_1}\cdot \vtau_1} \textcolor{teal}{ + \lambda_2 \cdot \vf_{\vtau_2}\cdot \vtau_2}}{{\color{black} \lambda_0 \cdot \vf_{\vthetaPRE}} \textcolor{brown}{+ \lambda_1 \cdot \vf_{\vtau_1}}\textcolor{teal}{+ \lambda_2 \cdot \vf_{\vtau_2}}}$
\end{algorithmic}
\caption{Pseudocode for \textcolor{brown}{removing hallucinations} and \textcolor{teal}{promoting abstraction} with \textcolor{black}{EWR}. Note that we apply $(\cdot)^2$ element-wise.}
\label{alg:ewr}
\end{algorithm}

In our proposed method, Elastic Weight Removal (EWR), we use the Fisher to combine models and task vectors with importance-weighted scaling factors for each parameter. Thereby, we aim to preserve positive behaviour in the model fine-tuned for dialogue response generation while removing the most important parameters in the anti-expert task vector, which induce hallucinated and extractive generation. 

We start by taking \cref{eq:param_combination} and setting $\vlambda_0$, which scales pre-trained parameters $\vthetaPRE$, to $\lambda_0 \cdot \vf_{\vtheta}$. On the other hand, note that $\lambda_0$ is equal to $1$ in \cref{eq:task_arithmetic} for task arithmetic.
Similarly, for each task vector $\vtau$, we replace the \emph{scalar} factor $\lambda_i$ with $\lambda_i \cdot \vf_{\vtau_i}$.
That is, we can still control the influence of each \emph{model} with a scalar hyper-parameter, while the diagonal observed Fisher from \cref{eq:diagfisher} controls \textit{individual parameters}.
Since the entries in $\vf$ can be orders of magnitudes smaller than the entries in $\vtheta$, we set the scaling constant $Z$ to be the sum of the products of scaling factors and their corresponding Fishers.
Therefore, our general parameter combination scheme is defined as:
\begin{align}
    \label{eq:ewr}
    \vthetaNEW &=
    \frac{\lambda_0 \cdot \vf_{\vthetaPRE} \cdot \vthetaPRE + \sum_{i=1}^N \lambda_i \cdot \vf_{\vtau_i} \cdot \vtau_i } {\lambda_0 \cdot \vf_{\vthetaPRE} + \sum_{i=1}^N \lambda_i \cdot \vf_{\vtau_i}} \text{,}
\end{align}
where again the sign of $\lambda_i$ indicates whether we add or negate behaviour.

Specifically, as illustrated in \cref{alg:ewr}, we obtain the initial parameters $\vthetaPRE$ by fine-tuning a pre-trained model on the full dataset $\mathcal{D}$. We then apply EWR to remove multiple undesirable behaviours, namely hallucinations and extractiveness. We first create an anti-expert task vector for hallucinations $\vtau_1$ from data $\mathcal{D}^\text{AE}$ and an expert for abstractiveness $\vtau_2$ from data $\mathcal{D}^\text{E}$. Then, we calculate the diagonal observed
Fisher of these three arrays of parameters. 
Finally, we combine them in accordance to Equation \ref{eq:ewr}.

To gain a better insight into EWR, let us restrict ourselves to the simpler case where we only subtract the task vector for hallucinations. In this case, we can rewrite our model combination as follows: 

\begin{align*}
    \vthetaNEW &=
    \frac{\lambda_0 \cdot \vf_{\vthetaPRE} \cdot \vthetaPRE - \lambda_1 \cdot \vf_{\vtau_1} \cdot \vtau_1} {\lambda_0 \cdot \vf_{\vthetaPRE} + \lambda_1 \cdot \vf_{\vtau_1}} \\
    &= \vthetaPRE - \frac{\lambda_1 \cdot \vf_{\vtau_1}} {\lambda_0 \cdot {\vf_{\vthetaPRE} + \lambda_1 \cdot \vf_{\vtau_1}}}\vthetaAE 
\end{align*}
This allows for the interpretation, that $\vf_\vthetaPRE$ and $\vf_{\tau_1}$ `compete' for how much each parameter should be changed --
parameters with large $\vf_\vthetaPRE$ are preserved, while others with $\vf_{\tau_1}$ are changed more significantly, as they contribute more to negative behaviour.
\subsection{(Anti-)Expert Data Selection}
\label{sec:expert_anti_expert}
As part of our method, we propose to select the examples to train (anti-)experts automatically via appropriate metrics. To create a dataset of hallucinated examples $\mathcal{D}^\text{AE}$, we resort to different strategies. For Wizard-of-Wikipedia (WoW), we make use of the annotations provided in the FaithDial \cite{dziri2022faithdial} dataset, where a subset of WoW was annotated according to the BEGIN \cite{dziri-etal-2022-evaluating} taxonomy. Under this taxonomy, responses whose information could not be inferred solely based on the grounding information, such as personal opinions, were marked as hallucinations.

In other datasets, such ground-truth annotations do not exist. Hence, we try different data augmentation techniques to artificially create hallucinated data.
We find that simply switching out the grounding information $\hat{\mathcal{K}}$ to one from a randomly sampled example from the dataset performs surprisingly well, similar to using ground-truth data. This may be explained by the fact that this forces the model to hallucinate, as the grounding information tends to be irrelevant to the response in such cases.
For methods like CaPE and DExpert (see \cref{sec:baselines}), which require a faithfulness expert in addition to a hallucination anti-expert, we use responses that are assigned an \emph{entailment} token when training CTRL.%

To create a dataset of abstractive behaviour $\mathcal{D}^\text{E}$, instead, we use the density and coverage metrics introduced in \citet{grusky2018newsroom}.
While coverage measures the ratio of unigrams from the grounding documents appearing in the response, density measures the average length of copied text spans.
First, we split the dataset examples into three equally-sized buckets indicating low, medium and high density.
Afterwards, we retain only high-coverage examples from the low-density subset, as they are less likely to contain hallucinations.
Nevertheless, this also decreases the size of the training data, which might have adverse effects.

\section{Experiments}
We experiment on multiple datasets outlined in \cref{sec:datasets} in order to compare EWR to state-of-the-art approaches for unlearning undesired behaviour, which we adapt to faithful dialogue generation. In addition to CaPE and task arithmetic described in \cref{sec:background}, we list a series of further baselines in \cref{sec:baselines}. Crucially, EWR can be deployed independently as well as \textit{on top of} several of them.

All experiments are implemented using the Huggingface transformers library \cite{wolf2020transformers} and all models are initialised with publicly available Flan-T5 checkpoints \cite{longpre2023flan}, which we have found to perform substantially better than previously introduced encoder-decoder models like BART \cite{lewis2020bart} or T5 \cite{2020t5}.
We organise our experiments using the Sisyphus \cite{peter2018sisyphus} workflow manager and release config files to reproduce our results.
All baseline models, with the exception of Quark which we train for 5 epochs, are trained for 10 epochs using an initial learning rate of $6.25e{-5}$, linear learning rate decay without warmup, and a batch size of 32.
We evaluate the models after each epoch on a held-out validation set and pick the model with the smallest validation loss.
On the other hand, the expert and anti-expert models are only trained for 5 epochs to not diverge too far from initialisation: see \cref{sec:expert_anti_expert} for details on their training procedures.
We use beam search with a beam size of 10 for decoding.

\subsection{Datasets}
\label{sec:datasets}
We evaluate EWR on Wizard-of-Wikipedia \cite[WoW]{dinan2018wizard}, an open-domain dataset for information-seeking dialogue where turns are grounded on Wikipedia snippets, and which contains a \emph{seen} and an \emph{unseen} split.
Furthermore, we rely on the DSTC9 \cite{kim2020beyond} extension of MultiWoZ 2.1 \cite{eric2019multiwoz}, where the original dialogues were augmented by turns that require grounding not only on structured databases but also on unstructured short FAQ documents. This dataset allows us to evaluate task-oriented settings where the existence of hallucinations may prove fatal, as users receive incorrect information.
Finally, we run experiments on DSTC11\footnote{\url{https://dstc11.dstc.community}}, a further extension of DSTC9 with customer reviews indicating subjective knowledge, to evaluate multi-domain settings, and FaithDial \cite{dziri2022faithdial}, which contains a de-hallucinated subset of WoW, for ablations.

\subsection{Baselines}
\label{sec:baselines}
\paragraph{CTRL}
\cite{keskar2019ctrl} introduces a sequence of control tokens $\mathbf{c}$ that are used to steer the model towards desirable behaviour: \begin{equation}
p(u_{T+1} \mid u_1^T, \hat{\mathcal{K}}, \mathbf{c}) \label{eq:ctrl}
\end{equation}
\citet{rashkin-etal-2021-increasing} adapt the model in \cref{eq:ctrl} to document-grounded dialogue by introducing \emph{entailment}, \emph{lexical overlap} and \emph{first-person} tokens, of which we employ the first two.
Entailment indicates whether the response is entailed by the documents, judged by an MNLI model, and lexical overlap splits the responses into three buckets according to low, medium, and high lexical overlap.
At training time, CTRL is trained on examples from all three buckets as well as both entailment and non-entailment examples.
At inference time, the generation is conditioned on tokens for entailment and high-overlap to promote faithfulness.

\begin{table*}[t!]
\centering
    \resizebox{\textwidth}{!}{\begin{tabular}{|l|lllllll|lllllll|}
    \hline
         & \multicolumn{7}{c|}{WoW$_{\text{seen}}$} & \multicolumn{7}{c|}{DSTC9}
         \\
         & BLEU($\uparrow$) & BERT F1($\uparrow$) & Critic($\downarrow$) & $Q^2 (\uparrow)$ & BERT F1($\uparrow$) &  F1($\uparrow$) & Density($\downarrow$) & BLEU($\uparrow$) & BERT F1($\uparrow$) & Critic($\downarrow$) & $Q^2 (\uparrow)$ & BERT F1($\uparrow$) &  F1($\uparrow$) & Density($\downarrow$)\\
         Model & \multicolumn{2}{c|}{$(y, \hat{y})$} & \multicolumn{5}{c|}{$(y, \hat{\mathcal{K}})$} & \multicolumn{2}{c|}{$(y, \hat{y})$} & \multicolumn{5}{c|}{$(y, \hat{\mathcal{K}})$}\\ \hline
Flan-T5$_{\text{base}}$ & 18.5 & \textbf{65.2} & 24.3 & 76.2 & 84.4 & 78.6 & 12.4 & 18.5 & 72.2 & 6.2 & 62.3 & 61.3 & 45.2 & 1.73\\ 
\, + TA  & 19.1 & 64.9 & 19.4 & 75.9 & 82.2 & 74.4 & 11.1  & 18.5 & 72.1 & 2.5 & 79.6 & 63.6 & 53.9 & 2.80 \\
\rowcolor{lightgray!20}
\, + EWR  & 18.1 \degrad{-0.4} & 64.4 \degrad{-0.8} & 18.1 \improv{-6.2} & 78.0 \improvUP{1.8} & 86.2 \improvUP{1.8} & 80.8 \improvUP{2.2} & 13.5 \degradUP{1.1}& \textbf{20.0} \improvUP{1.5} & 72.3 \improvUP{0.1} & 4.3 \improv{-1.9} & 78.4 \improvUP{16.1} & 64.4 \improvUP{3.1} & 55.6 \improvUP{10.4} & 3.22 \degradUP{1.49}\\ 
CaPE & 18.8 & 64.8 & 13.2 & 78.2 & 83.7 & 75.9 & 11.2 & 17.3 & 71.8 & 2.3 & 72.5 & 63.3 & 52.6 & 2.63\\ 
\rowcolor{lightgray!20}
\, + EWR  & 19.0  \improvUP{0.2}& 64.3  \degrad{-0.5}& 9.4  \improv{-3.8} & 78.7 \improvUP{0.5} & 88.2  \improvUP{4.5}& 83.0  \improvUP{7.1}& 13.6  \degradUP{2.4}& 16.7  \degrad{-0.6} & 71.9  \improvUP{0.1}& 2.6  \degradUP{0.3} & 79.2 \improvUP{6.7} & 64.3  \improvUP{1.0}& 54.0 \improvUP{1.4} & 2.76 \degradUP{0.13}\\ 
CTRL  & \textbf{19.5} & 64.8 & 10.3 & 83.9 & 87.8 & 82.3 & 13.9 & 17.6 & 71.8 & 5.3 & 79.8 & 64.5 & 57.8 & 3.30\\ 
\, + TA  & 19.3 & 64.7 & 8.9 & 82.7 & 87.0 & 81.2 & 13.0 & 18.0 & 71.9 & 1.2 & \textbf{89.5} & \textbf{66.5}&\textbf{63.6} & 4.53 \\ 
\rowcolor{lightgray!20}
\, + EWR  & 18.4 \degrad{-0.8} & 63.7 \degrad{-1.1}& \textbf{5.7} \improv{-4.6} & 86.8 \improvUP{2.9} & 91.3 \improvUP{3.5} & 87.7 \improvUP{5.4} & 16.3  \degradUP{2.4}& 19.4 \improvUP{1.7} & 72.3 \improvUP{0.5} & 2.3 \improv{-3.0} & 85.3 \improvUP{5.5} & 65.5 \improvUP{1.0} & 60.6 \improvUP{2.8} & 3.80 \degradUP{0.5}\\ \hline
DExperts & 18.0 & 64.3 & 14.8 & 79.6 & 87.0 & 82.2 & 14.3 & 17.1 & 71.5 & 2.9 & 74.9 & 63.6 & 55.7 & 2.83\\ 
Quark & 17.2 & 63.6 & 7.9 & \textbf{91.9} & \textbf{92.6} & \textbf{90.2} & 18.6 & 19.0 & 72.5 & 5.7 & 73.1 & 62.7 & 49.8 & 2.03\\ 
Noisy Channel & 18.4 & 64.8 & 24.0 & 78.6 & 85.0 & 79.8 & 13.1 & 18.6 & 72.5 & 5.1 & 67.1 & 62.7 & 48.4 & 2.18\\
\hline
\hline
         Flan-T5$_{\text{large}}$  & 18.6 & \textbf{65.5} & 26.7 & 77.8 & 83.8 & 77.5 & 12.3 & 18.6 & 72.2 & 6.9 & 64.0 & 61.2 & 44.7 & 1.81\\ 
\, + TA  & 19.1 & 65.1 & 16.7 & 80.2 & 84.6 & 77.8 & 12.6 & 19.0 & 72.6 & 3.7 & 74.3 & 64.4 & 55.6 & 3.50 \\ 
\rowcolor{lightgray!20}
\, + EWR  & 17.3  \degrad{-1.3}& 64.2  \degrad{-1.3} & 16.9  \improv{-9.8}& 80.3 \improvUP{2.5} & 88.3  \improvUP{4.5}& 83.9  \improvUP{6.4}& 14.9  \degradUP{2.6}& 19.1  \improvUP{0.5}&\textbf{72.8}\improvUP{0.6}& 2.8  \improv{-4.1}& 83.8 \improvUP{19.8} & 64.8  \improvUP{3.6}& 57.3 \improvUP{12.6} & 3.48 \degradUP{1.67}\\ 
CaPE  & 19.0 & 65.1 & 13.0 & 79.5 & 83.7 & 75.4 & 11.3 & 17.2 & 72.2 & 4.3 & 73.3 & 64.4 & 53.2& 2.82\\ 
\rowcolor{lightgray!20}
\, + EWR  & 18.2  \degrad{-0.8}& 64.1  \degrad{-1.0}& 9.3  \improv{-3.7}& 80.4 \improvUP{0.9} & 89.4  \improvUP{5.7}& 84.9  \improvUP{9.5}& 15.2  \degradUP{3.9}& 16.2  \degrad{-1.0}& 71.6  \degrad{-0.6}& 1.1  \improv{-3.2}& 74.9 \improvUP{1.6} & 64.1  \degrad{-0.3}& 54.1 \improvUP{0.9} & 3.00 \degradUP{0.18}\\ 
CTRL  & \textbf{19.8} & 65.2 & 11.3 & 82.0 & 87.3 & 81.5 & 13.4 & \textbf{19.5} & \textbf{72.8} & 6.8 & 77.4 & 63.8 & 52.7 & 2.73\\ 
\, + TA  & 19.2 & 64.9 & 7.2 & 84.3 & 86.8 & 80.6 & 13.0 & 19.3 & 72.7 & 2.6 & 79.3 & \textbf{65.9} & 57.5 & 3.37 \\ 
\rowcolor{lightgray!20}
\, + EWR  & 18.6  \degrad{-1.2}& 64.3  \degrad{-0.9}& \textbf{7.0 } \improv{-4.3}& 85.8 \improvUP{5.4} & 90.5 \improvUP{3.2}& 86.8  \improvUP{5.3}& 16.8  \degradUP{3.4}& 18.1  \degrad{-1.4}& 72.7  \degrad{-0.1}&\textbf{0.8} \improv{-6.0}& \textbf{84.3} \improvUP{6.9} & 65.2  \improvUP{1.4}& \textbf{59.5} \improvUP{6.8} & 3.83 \degradUP{1.1}\\ \hline

DExperts & 18.3 & 64.7 & 17.9 & 79.8 & 81.7 & 71.4 & 12.7 & 18.2 & 72.2 & 4.2 & 70.5 & 63.9 & 54.9 & 2.78\\ 
Quark & 18.0 & 64.0 & 9.1 & \textbf{91.4} & \textbf{91.2} & \textbf{88.1} & 16.9 & 20.3 & 73.3 & 6.0 & 74.7 & 64.9 & 54.3 & 3.09 \\ 
Noisy Channel & 18.8 & 65.1 & 22.3 & 77.2 & 85.5 & 80.2 & 13.3 & 18.4 & 72.3 & 6.1 & 67.2 & 62.2 & 47.4 & 2.20\\ \hline
    \end{tabular}}
    \caption{Main results on the WoW$_\text{seen}$ and DSTC9 datasets indicating: i) performance in dialogue generation comparing true $\hat{y}$ and predicted $y$ responses (BLEU and BERT F1); ii) faithfulness between predicted response $\hat{y}$ and gold-truth knowledge $\hat{\mathcal{K}}$ (Critic, $Q^2$, BERT F1, F1); 3) abstractiveness (Density). We report several baselines adapted for faithful generation and show how Task Arithmetic (TA) and Elastic Weight Removal (EWR, ours) can be deployed on top of vanilla pre-trained models, like Flan-T5, or on top of other methods like CaPE and CTRL. We indicate relative improvements in green and relative degradations in red with a down arrow.
    }
    \label{tab:main_results}
    
\end{table*}

\paragraph{Quark}
\cite{lu2022quark} provides a way of unlearning undesirable behaviour by repeatedly sampling from the model after each epoch, quantising the utility of each sample, and then conditioning on a reward token according to the quantised utility in training.
When decoding, the desirable tokens are used to condition the model, analogously to CTRL.
Noting this similarity, we therefore employ the same tokens as in CTRL, which allows for a direct comparison between these methods.

\paragraph{DExperts}
\cite{liu2021dexperts}, akin to our methods, makes use of an expert and anti-expert model in order to steer generations towards desirable behaviour.
However, instead of combining models in \emph{parameter space}, the models are combined at inference time by weighting probabilities with the ratio of the densities of the expert and anti-expert: \begin{align}
    &p(u_{T+1} \mid u_1^T, \hat{\mathcal{K}}) \propto \\
    &\quad p_\vthetaFT(u_{T+1} \mid u_1^T, \hat{\mathcal{K}})\cdot\frac{p_\vthetaE(u_{T+1} \mid u_1^T, \hat{\mathcal{K}})}{p_{\vthetaAE}(u_{T+1} \mid u_1^T, \hat{\mathcal{K}})} \text{.} \nonumber
\end{align}
Intuitively, tokens with high expert probability are encouraged, whereas tokens with high anti-expert probability are discouraged.

We use the same expert and anti-expert models as in CaPE to fairly compare both methods.

\paragraph{Noisy Channel Model}
\citep{daheim-etal-2022-controllable} introduces a noisy channel model for document-grounded dialogue, where the model is split into two factors: \begin{align}
p(\hat{\mathcal{K}} \mid u_1^T, u_{T+1}) \cdot p(u_{T+1} \mid u_1^T) \text{,}
\end{align}
where both components can be seen as a faithfulness and fluency expert, respectively.
We use their reranking method to rescore generations obtained from our baseline models.

\subsection{Metrics}
\label{sec:metrics}
We measure the lexical similarity of the generated responses with their ground-truth annotations through the sacrebleu \cite{post2018sbleu} implementation of BLEU \cite{papineni2002bleu}.
Furthermore, we score the semantic similarity of the generated response with both the ground-truth response and the grounding documents through BERTScore \cite{Zhang2020BERTScore}.\footnote{We make use of the \emph{deberta-large-mnli} checkpoint.}
Additionally, we employ the hallucination critic introduced by \citet{dziri2022faithdial}\footnote{ \url{https://huggingface.co/McGill-NLP/roberta-large-faithcritic}.}, a re-implementation of the QA-QG-based $Q^2$ metric \cite{honovich-etal-2021-evaluating}, as well as token-level F1 to further evaluate faithfulness.

To measure abstractiveness, we use the \emph{Density} metric \cite{grusky2018newsroom}, which indicates the average squared length of extractive snippets from $\hat{\mathcal{K}}$ in $u_{T+1}$ such that a lower density indicates less copying.

\begin{figure*}[t!]
\begin{subfigure}{.4\textwidth}
\centering
\begin{tikzpicture}
\begin{axis}[
    xlabel=Scaling Factor (Abstraction Expert),
    xmin=0, xmax=0.15,
    ymin=0, ymax=40,
    xtick={0.0,0.05,...,0.15},
    ytick={0,5,...,100},
    style=ultra thick,
    legend cell align={left},
    legend style={at={(0.025,0.975)}, anchor=north west}
            ]
\addplot[smooth,mark=x,red, mark options={mark size=2pt, line width=2pt}] plot coordinates {
    (0.0, 18.7)
    (0.025, 18.4)
    (0.05, 18.8)
    (0.075, 20.1)
    (0.1, 21.8)
    (0.125, 23.5)
    (0.15, 27.0)
};
\addplot[smooth,mark=*,orange] plot coordinates {
    (0.0, 13.5)
    (0.025, 13.3)
    (0.05, 12.8)
    (0.075, 12.1)
    (0.1, 11.4)
    (0.125, 10.5)
    (0.15, 7.3)
};
\draw [dashed, red] (0,243) -- (150,243);
\draw [name path=right, gray] (131,0) -- (131,500);
\draw [dashed, orange] (0,124) -- (150,124);
\draw [name path=left, gray] (61,0) -- (61,500);
       \addplot [gray!15] fill between[
            of=left and right,
            reverse=true,
        ];

\addlegendentry{Critic}
\addlegendentry{Density}
\end{axis}
\end{tikzpicture}
\subcaption{Faithfulness--Abstractiveness Trade-Off}
\label{fig:abstractiveness_expert}
\end{subfigure}
\hspace{2cm}
\begin{subfigure}{.4\textwidth}
\begin{tikzpicture}
\begin{axis}[
    xlabel=Scaling Factor (Hallucination Anti-Expert),
    xmin=0, xmax=0.2,
    ymin=0, ymax=40,
    xtick={0.0,0.05,...,0.2},
    ytick={0,5,...,100},
        legend cell align={left},
    legend style={at={(0.025,0.975)}, anchor=north west, style=ultra thick}
            ]
\addplot[smooth,mark=x,red, mark options={mark size=2pt, line width=2pt}] plot coordinates {
    (0.0, 24.3)
    (0.025, 23.3)
    (0.05, 22.6)
    (0.075, 21.0)
    (0.1, 20.2)
    (0.125, 19.2)
    (0.15, 18.1)
    (0.175, 18.1)
    (0.2, 18.0)
};
\addplot[smooth,mark=*,orange,mark options={mark size=2pt, line width=2pt}] plot coordinates {
    (0.0, 12.4)
    (0.025, 12.8)
    (0.05, 13.1)
    (0.075, 13.3)
    (0.1, 13.6)
    (0.125, 13.6)
    (0.15, 13.5)
    (0.175, 13.4)
    (0.2, 13.4)
};
\addplot[smooth,mark=+,Cerulean, mark options={mark size=2pt, line width=2pt}] plot coordinates {
    (0.0, 18.5)
    (0.025, 18.5)
    (0.05, 18.5)
    (0.075, 18.4)
    (0.1, 18.3)
    (0.125, 18.2)
    (0.15, 18.1)
    (0.175, 18.1)
    (0.2, 17.7)
};
\draw [dashed, red] (0,243) -- (250,243);
\draw [dashed, Cerulean] (0,185) -- (250,185);
\draw [dashed, orange] (0,124) -- (250,124);
\addlegendentry{Critic}
\addlegendentry{Density}
\addlegendentry{BLEU}
\end{axis}
\end{tikzpicture}
\subcaption{Faithfulness--Performance Trade-Off}
\label{fig:controllability}
\end{subfigure}
\caption{Metrics for EWR on top of Flan-T5$_\text{base}$ on the seen test split of WoW. (a) Varying the influence of the abstraction expert model gives control over the trade-off between faithfulness and abstractiveness, when fixing the scaling factor of the hallucination anti-expert. Dashed lines indicate baseline Flan-T5$_\text{base}$ performance and grey shading improvements over it in terms of both criteria. (b) Using only the hallucination anti-expert, varying the scaling factor again shows a decrease in Critic score but also an increase in Density and a slight decline in BLEU.}
\label{fig:merged}
\end{figure*}
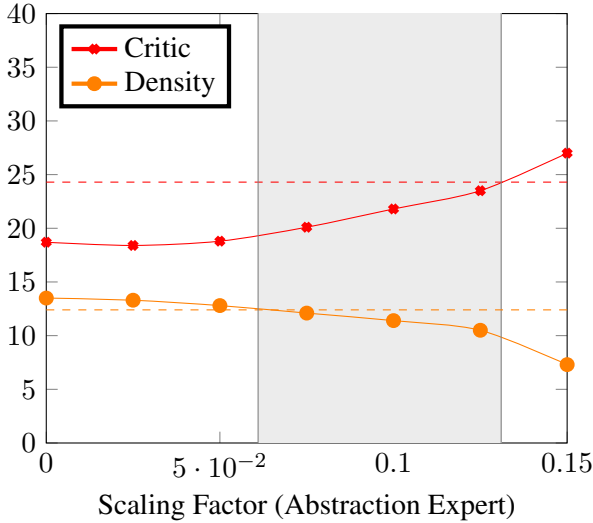
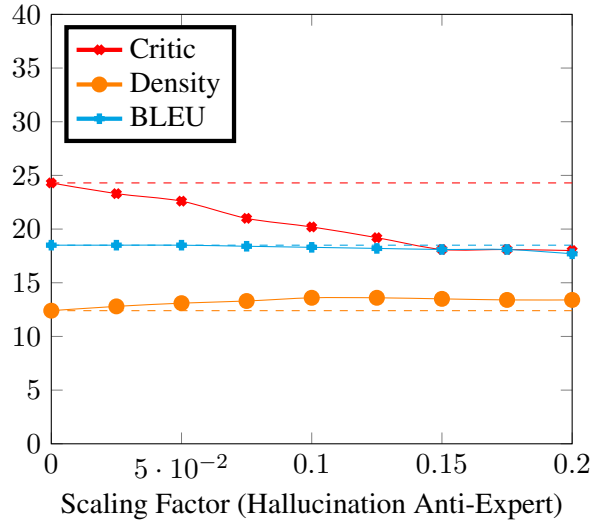

\section{Results}
\label{sec:results}
This section is structured as follows.
We first introduce our main results on Wizard-of-Wikipedia and DSTC9 in Section \ref{sec:main_results}.
Then, we characterise trade-offs between faithfulness and abstractiveness in Section \ref{sec:tradeoff} before discussing the controllability of model interpolation in Section \ref{sec:controllability}.
Finally, we discuss ablations on various datasets, such as multi-document settings, in Section \ref{sec:ablations}. Later on, in \cref{sec:human_evaluation} we report human evaluation results.

\subsection{Main Results on Faithfulness}
\label{sec:main_results}

We start by focusing on results for de-hallucinated models in \cref{tab:main_results}. These confirm the intuition that using anti-experts to construct negative parameter sets for hallucination and subtracting them from dialogue models can indeed improve faithfulness at minimal degradation in other metrics.
We observe that EWR generally increases faithfulness according to all metrics when applied to various base models, sometimes by a large margin.
Furthermore, EWR tends to yield more significant increases than Task Arithmetic, which in some metrics, especially BERT and token-level F1, does not outperform its corresponding base models.
Using EWR on top of CTRL, we obtain state-of-the-art results in terms of faithfulness and sometimes are even able to outperform strong baselines like Quark.

While an additional faithfulness expert in CaPE generally improves over using only an anti-expert, we observe fast degradation in terms of BLEU and BertScore on DSTC9. This stems from the comparatively small size of the expert training set after partitioning the dataset.

Moreover, all baselines that we adapt to promote faithfulness indeed hallucinate less according to automatic metrics.
CTRL and Quark perform strongly, confirming the effectiveness of control tokens and \emph{iteratively} applying them by sampling from the model during training.
Inference-time model combination with both DExperts and noisy channel reranking is mostly outperformed by EWR, Task Arithmetic, and CaPE, except for Flan-T5$_\text{base}$ on WoW.
This is significant, as our models also incur no overhead at inference time and have fewer parameters (DExperts and the noisy channel reranking triple their number).
Nevertheless, we need to note that the performance of the noisy channel model increases with its beam size, which is kept identical for all methods in our experiments for comparability.

All methods improve faithfulness across model sizes, using both the base Flan-T5 model with 250M parameters and the large model with 780M parameters. Contrary to our expectations, however, the larger model often does not improve over the smaller counterpart notwithstanding it boasts thrice the parameter count.

Furthermore, our trends differ slightly across datasets.
The gains of CTRL and Quark are much more conspicuous in WoW than DSTC9.
We attribute this to the fact that in DSTC9, the ground-truth documents contain FAQs. In these cases, the question might not be as important for the control tokens. Gold responses contain follow-up questions at every turn, as the system simulates a customer service agent. This might  decrease the effectiveness of tokens, especially for lexical overlap and might also affect automatic metrics.
This hints at the potential for future work to devise better methods of creating control tokens.

Nevertheless, our results in \cref{tab:main_results} also illustrate that increased faithfulness comes at the cost of increased extractiveness, as described by the Density metric. We investigate this phenomenon further in the following subsection.

\subsection{Faithfulness--Abstractiveness Trade-Off}
\label{sec:tradeoff}
As our main experiments show that the improvements in faithfulness using EWR incur an increase in extractiveness, we now outline experiments using an additional abstractiveness expert to reduce this effect.
The results are highlighted in \ref{fig:merged} \subref{fig:abstractiveness_expert} when fixing the influence of the hallucination anti-expert and varying the one of the abstractiveness expert on WoW using a Flan-T5$_\text{base}$ model.
From the plot, it emerges that we can indeed control the \emph{trade-off} between faithfulness and abstractiveness and improve over the baseline in both dimensions, in the interval indicated by the greyed area.

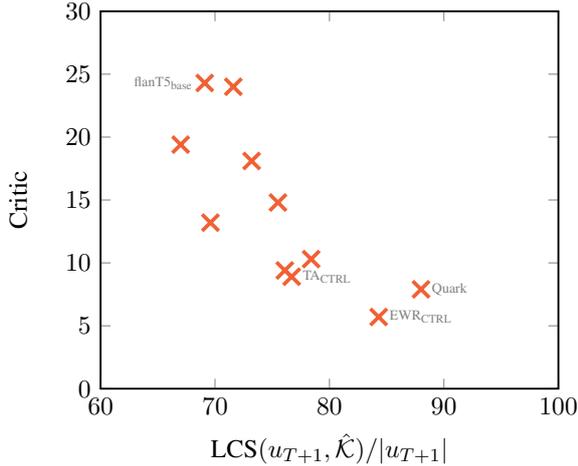
\begin{figure}[t]
\centering
\resizebox{\columnwidth}{!}{\begin{tikzpicture}
\begin{axis}[
    xlabel={LCS$(u_{T+1}, \hat{\mathcal{K}})/|u_{T+1}|$},
    ylabel=Critic,
    xmin=60, xmax=100,
    ymin=0, ymax=30,
    xtick={0.0,10, ..., 100},
    ytick={0,5,...,30},
    style=thick,
    legend style={at={(0.025,0.975)}, anchor=north west}
            ]
\addplot+[ultra thick,only marks,mark=x,RedOrange, mark size=5pt] plot coordinates {
(69.1, 24.3) %
(71.6, 24.0) %
(67.0, 19.4)
(73.2, 18.1)
(69.6, 13.2)
(76.1, 9.4)
(78.4, 10.3) %
(76.7, 8.9) %
(84.3, 5.7) %
(75.5, 14.8) %
(88.0, 7.9) %
};
\node[draw=RedOrange!0, thick] at (305, 79) {\tiny \color{gray} Quark};
\node[draw=RedOrange!0, thick] at (280, 57) {\tiny \color{gray} $\mathbf{\text{EWR}_{\text{CTRL}}}$};
\node[draw=RedOrange!0, thick] at (198, 89) {\tiny \color{gray} $\mathbf{\text{TA}_{\text{CTRL}}}$};
\node[draw=RedOrange!0, thick] at (55, 243) {\tiny \color{gray} $\mathbf{\text{flanT5}_{\text{base}}}$};

\end{axis}
\end{tikzpicture}}
\caption{Improvements in faithfulness (measured by Critic) tend to incur an increase in extractiveness (measured by LCS) on WoW.}
\label{fig:abstractiveness_faithfulness}
\end{figure}

To further quantify this trade-off, which has also been described in related works \cite{daheim-etal-2022-controllable, aksitov2023characterizing, dziri2022faithdial}, we make use of the ratio of the length of the longest common subsequence between $u_{T+1}$ and $\hat{\mathcal{K}}$ and the length of $u_{T+1}$ (LCS), which corresponds to the precision term in ROUGE-L \cite{lin2004rouge}.
We plot this dependency in Figure \ref{fig:abstractiveness_faithfulness} for models based on Flan-T5$_\text{base}$ on the seen split of WoW.
Comparing LCS against the Critic metric, there is a clear trend towards more extractiveness with increased faithfulness.
Nevertheless, a better Critic score does not always imply an increase in LCS. For example, EWR$_\text{CTRL}$ outperforms Quark in terms of both metrics in this experiment.

\subsection{Scaling Factors \& Controllability}
\label{sec:controllability}
Next, we assess how much control EWR gives over faithfulness scores within an acceptable range of BLEU, which measures performance.
\Cref{fig:merged} \subref{fig:controllability} highlights that there is indeed a large region of factors along which faithfulness constantly improves within a narrow range of BLEU scores, supporting the ideas of connected low loss regions \cite{ainsworth2023git}.
However, density increases with faithfulness, which corresponds to the trade-off we identified previously in \cref{sec:tradeoff}.

\subsection{Generalisation to Additional Datasets}
\label{sec:ablations}
\begin{table}
    \centering
    \resizebox{\columnwidth}{!}{\begin{tabular}{|l|llllll|}
    \hline
         &  BLEU($\uparrow$) & BERT F1($\uparrow$) & Critic($\downarrow$) & $Q^2 (\uparrow)$ & BERT F1($\uparrow$) &  F1($\uparrow$) \\
         Model & \multicolumn{2}{c|}{$(y, \hat{y})$} & \multicolumn{4}{c|}{$(y, \hat{\mathcal{K}})$} \\ \hline 
         & \multicolumn{6}{c|}{WoW$_{\text{unseen}}$} \\
Flan-T5$_{\text{base}}$ & 18.1 & 65.1 & 22.7 & 74.0 & 84.8 & 78.7\\ 
\; + Task Arithmetic & 18.8 & 64.7 & 19.2 & 75.7 & 82.8 & 75.0 \\ 
\rowcolor{lightgray!20}
\; + EWR & 17.4  \degrad{-0.7}& 64.4  \degrad{-0.7}& 17.7  \improv{-5.0}& 78.4 \improvUP{4.4} &86.9  \improvUP{2.1}& 81.6 \improvUP{2.9}\\
         & \multicolumn{6}{c|}{DSTC11} \\
Flan-T5$_{\text{base}}$ & 7.9 & 62.4 & 76.6 & 49.7 & 54.6 & 37.1\\ 
\; + Task Arithmetic & 8.0 & 62.5 & 60.0 & 51.0 & 59.9 & \textbf{43.6}\\ 
\rowcolor{lightgray!20}
\; + EWR &\textbf{9.6} \improvUP{1.7} & \textbf{64.2} \improvUP{1.8} &\textbf{41.1} \improv{35.5} & \textbf{57.3} \improvUP{7.6} &\textbf{60.0} \improvUP{5.4} & 38.6 \improvUP{1.5} \\ 
         & \multicolumn{6}{c|}{FaithDial} \\
Flan-T5$_{\text{base}}$ & 15.1 & 69.6 & 0.3 &\textbf{66.4}  & 80.9 & 73.7\\ 
\; + Task Arithmetic &\textbf{15.3} & 69.5  &\textbf{0.1} & 57.5 & 77.3 & 67.6 \\ 
\rowcolor{lightgray!20}
\; + EWR & 14.9  \degrad{-0.2}&\textbf{70.1} \improvUP{0.5}&\textbf{0.1} \improv{-0.2}& \textbf{66.4} \nochange{0.0}& \textbf{81.7} \improvUP{0.8}&\textbf{75.0} \improvUP{1.3}\\ 
         \hline
    \end{tabular}}
    \caption{EWR also improves faithfulness in a range of other settings, namely on unseen topics (WoW$_\text{unseen}$), multi-document corpora (DSTC11), and datasets with cleaned ground-truth annotations (FaithDial).}
    \label{tab:dataset_ablations}
\end{table}
In this section, we aim to study the performance of EWR in challenging settings, namely on: i) unseen topics that require significant generalization (WoW unseen), ii) multi-document corpora (DSTC11), and iii) cleaned training and test data that does not contain hallucinations in ground-truth annotations (FaithDial). We report the results in \cref{tab:dataset_ablations}.

In summary, we observe the following:
1) EWR shows improvements over all settings, especially in terms of generalizing to unseen topics and in a multi-document setting, indicating that it is robust and generalises to multiple domains. Furthermore, we can even improve faithfulness metrics when training and evaluating on the cleaned FaithDial dataset.
2) Task Arithmetic can improve results on a multi-document setting and some metrics on the unseen set but fails to improve BERT F1 and F1 on WoW unseen and FaithDial.

\section{Human Evaluation}
\label{sec:human_evaluation}
In addition to the automatic evaluation, we conduct a human evaluation on both WoW and DSTC9 with the help of three expert annotators \footnote{All annotators are graduate students in NLP.},
using the Attributable to Interpretable Source (AIS) framework \cite{rashkin2021measuring}. First, we ask them to score responses as attributable (A) only if all their content can be attributed to an interpretable source in the given context, i.e., the knowledge that grounds the dialogue response.
Furthermore, we ask annotators to rate cooperativeness (C), i.e. the ability of the model to connect with and follow up on user turns,
on a 3-point Likert scale. Here, 1 indicates a response that does not cooperate with the dialogue, 2 a response that brings the dialogue forward, and 3 a response that acknowledges the previous utterances and responds with a follow-up question.
Lastly, annotators rate paraphrasing (P) on a binary scale, where 2 indicates that the knowledge is paraphrased non-trivially and 1 indicates substantial copying.

\begin{table}[ht]
    \centering
    \resizebox{\columnwidth}{!}{\begin{tabular}{|l|lll|lll|}
    \hline
    Model & \multicolumn{3}{c|}{WoW} & \multicolumn{3}{c|}{DSTC9} \\
    & A ($\uparrow$) & C ($\uparrow$) & P ($\uparrow$)& A ($\uparrow$) & C ($\uparrow$) & P ($\uparrow$) \\
    \hline
        Flan-T5$_\text{base}$ & 72.3 & 1.74 & 1.19 & 89.7 & 2.83 & 1.71 \\
        \rowcolor{lightgray!20} EWR$_\text{abs}$ & 75.1 & 1.62 & 1.25 & 94.7$^\ast$ & 2.41 & 1.49 \\
        CTRL & 85.5$^\ast$ & 1.58 & 1.12 & 94.7$^\ast$ & 2.72 & 1.42 \\
        TA$_\text{CTRL}$ & 88.8$^\ast$ & 1.58 & 1.16 & 97.0$^\ast$ & 2.63 & 1.40 \\
        \rowcolor{lightgray!20} EWR$_\text{CTRL}$ & 96.8$^\dagger$ & 1.50 & 1.08 & 98.0$^\dagger$ & 2.50 & 1.36 \\
        Quark & 93.1$^\dagger$ & 1.51 & 1.05 & 86.0 & 2.89 & 1.66 \\ \hline
       \end{tabular}}
       \caption{Human evaluation on $218$ examples annotated by $3$ expert annotators each. We measure attributability (A), Co-cooperativeness (C), and the amount of paraphrasing (P). %
       $^\ast$ indicates significant improvements wrt.\ Flan-T5$_\text{base}$ and $^\dagger$ also wrt.\ to the next best method with $p < 0.05$.}
       \label{tab:human_eval}
\end{table}
Table \ref{tab:human_eval} shows the results obtained from the annotators for the A, C, and P categories with agreements of $0.61, 0.51, 0.53$, respectively, in terms of Fleiss'~$\kappa$.
Generally, we observe that human evaluation results for attributability (A) confirm the results based on automatic faithfulness metrics as they display similar patterns.
In particular, all methods improve over vanilla Flan-T5, with CTRL and Quark performing similarly on average and outperforming each other on the two different datasets.
Task Arithmetic and EWR give improvements over CTRL on both datasets, corroborating our intuition that subtracting anti-expert task vectors for hallucination improves faithfulness.
Most notably, EWR$_\text{CTRL}$ improves over all other methods, including Task Arithmetic and Quark, by a statistically significant margin in human evaluation.

However, our results also confirm the trade-off between faithfulness and both paraphrasing (which reflects abstractiveness) and cooperativeness. In fact, increased attributability leads to a decrease in both other criteria.
Nevertheless, we can conclude that EWR with an abstraction expert, labelled EWR$_\text{abs}$, improves both paraphrasing on WoW and attributability on both datasets compared to vanilla Flan-T5.
While EWR$_\text{abs}$ does not outperform this baseline in paraphrasing on DSTC9, we believe that this stems from the way the expert dataset $\mathcal{D}^\text{E}$ is constructed.
As the ground-truth responses in DSTC9 contain longer follow-up questions, it is likely that density-based binning does not pick up nuances, such as the difference between non-paraphrased responses and follow-up questions independent from the knowledge.
Future work might therefore attempt to devise better methods of data selection or augmentation for expert training.

\section{Related Work}

\paragraph{Hallucination in LMs}
The impressive abilities of LMs are offset by the potential for generating entirely false text, as they lack an understanding of what is factual and what is not \cite{ji2022survey, bang2023multitask, qin2023chatgpt, choi2023chatgpt, thoppilan2022lamda}. 
Consequently, there is an increasing interest in the NLP community to tackle the problem of hallucination in knowledge-grounded neural language generation \cite{ji2022survey}. This issue encompasses several tasks such as data-to-text generation \cite{wiseman-etal-2017-challenges, parikh-etal-2020-totto}, machine translation \cite{raunak-etal-2021-curious,wang-sennrich-2020-exposure}, summarisation \cite{durmus-etal-2020-feqa,kang-hashimoto-2020-improved}, generative question answering \cite{li2021addressing}, and dialogue generation \cite{dziri-etal-2021-neural, dziri-etal-2022-evaluating, rashkin-etal-2021-increasing, daheim-etal-2022-controllable}. These studies aim to address the issue of hallucination by either developing automatic metrics to detect it \citep{wiseman-etal-2017-challenges}, or by identifying potential causes such as out-of-domain generalisation, noisy training data, and exposure bias resulting from maximum likelihood estimation (MLE) training \citep{kang-hashimoto-2020-improved,raunak-etal-2021-curious,wang-sennrich-2020-exposure, dziri-etal-2021-neural}. 

\paragraph{Hallucination in Neural Dialogue Models}

The issue of hallucinations in knowledge-grounded neural dialogue generation has been studied intensively recently \cite{roller-etal-2021-recipes,shuster-etal-2021-retrieval-augmentation,dziri-etal-2021-neural, dziri2022faithdial,razumovskaia2022crossing, daheim-etal-2022-controllable}. Existing approaches mainly focus on addressing hallucinations by engineering loss functions or enforcing consistency constraints. For example, conditioning generation on control tokens \citep{rashkin-etal-2021-increasing}, learning a token-level hallucination critic to detect and replace problematic entities \citep{dziri-etal-2021-neural}, or incorporating a module to retrieve relevant knowledge \citep{shuster-etal-2021-retrieval-augmentation}.
However, they are susceptible to replicating or even amplifying the noise present in the training data. Recent research by \citet{dziri-etal-2022-origin} indicated that over 60\% of three commonly used dialogue benchmarks are prone to hallucination, which affects even by models designed to increase faithfulness and creativity. To address this issue, \citet{dziri2022faithdial} proposed a hallucination-free dialogue benchmark, where hallucinated responses were re-annotated.

\paragraph{Controllable text generation}
Previous works have examined various controllable techniques aimed primarily at minimising toxicity and sentiment-controlled generation. \citet{liu2021dexperts} proposed DExperts which involves combining a pre-trained language model with expert and anti-expert language models that respectively model text with desirable and undesirable attributes. 
\citet{lu2022quark} introduced an RL-based algorithm to unlearn misalignments by optimising a reward function that quantifies unwanted properties, while staying close to the original model. \citet{ilharco2023editing} proposed applying arithmetic operations on the model weights to unlearn negative properties. 
To ensure dialogues are faithful, other works borrow from the use of control-code-style input tokens, as seen in models like CTRL \cite{keskar2019ctrl, rashkin-etal-2021-increasing} and the LFT model \cite{10.1162/tacl_a_00027}. Although controllable generation has been previously used to improve qualities such as engagement in open-ended dialogue data \cite{see-etal-2019-makes}, our work is focused on knowledge-grounded dialogues with the aim of increasing response faithfulness and creativity. %

\paragraph{Fisher Information Matrix}
EWR differs from previous work using Fisher Information in various respects. In particular, contrary to Elastic Weight Consolidation \cite{kirkpatrick2017overcoming}, where the Fisher is used as a prior for regularisation during training, we only make use of the Fisher \emph{after} completing training to combine the pre-trained model with (anti-)experts.
We furthermore differ from Fisher-weighted model averaging \citep{matena2021merging} as we do not merge separate models trained on different tasks but rather \textit{task vectors}, which consist of changes between initialisation and fine-tuning and can be incapable of producing meaningful generations \textit{per se}. Moreover, we consider merging \textit{negative} experts similar to the task arithmetic framework.

\section{Conclusion \& Future Work}
In this paper, we introduced Elastic Weight Removal (EWR), a novel method for reducing undesirable behaviours and encouraging positive behaviours in conditional language generation.
In particular, we create vectors of differences between pre-trained models and (anti-)expert models fine-tuned on examples of negative or positive behaviours. These vectors of differences are subtracted from or added to pre-trained models, similar to task arithmetic, but also weighted according to their corresponding Fisher Information. 

We show how EWR reduces hallucinations in document-grounded dialogue response generation across multiple datasets, including information-seeking dialogue (WoW) and task-oriented dialogue (DSTC9). Based on automated metrics and human evaluation,
EWR improves faithfulness over multiple baselines, most notably task arithmetic. In addition, we compare EWR with a series of state-of-the-art methods for unlearning negative behaviours that we adapt for faithful dialogue response generation.
We find that EWR outperforms other methods like DExperts or Noisy Channel reranking and obtains complementary improvements on others, such as CTRL, to achieve results that are competitive with Quark.
Our ablations show that these improvements hold in other, challenging settings, where topics are unseen during training, the ground-truth knowledge consists of multiple documents, or the training data is already stripped of hallucinations.

Moreover, we note through extensive ablations that faithfulness comes at the expense of abstraction.
Therefore, we outline how an abstraction expert can be combined with the hallucination anti-expert to promote responses that are simultaneously more faithful \textit{and} abstractive than the baseline.

The significance of the present work, arguably, is that it outlines a previously unexplored way of promoting faithfulness in document-grounded dialogue by using experts and anti-experts not at inference time---and thereby incurring significant overhead---but rather to navigate the parameter space in order to obtain an improved array of parameters without altering the model architecture.

This opens up many potential areas for future work, such as evaluating EWR in other domains (such as retrieval-augmented models) or removing further dimensions (such as toxicity and redundancy) to afford better control over the generations of dialogue models.
Moreover, in the case of document-grounded dialogue, another line of research is developing more sophisticated data augmentation techniques to create data for expert and anti-expert training.

\section*{Acknowledgements}
This project has received funding by the German Federal Ministry of Education and Research and the Hessian Ministry of Higher Education, Research, Science and the Arts within their joint support of the National Research Center for Applied Cybersecurity ATHENE.

\bibliography{anthology,custom}
\bibliographystyle{acl_natbib}

\end{document}